\documentclass[preprint,12pt]{elsarticle}

\usepackage{lineno,hyperref}
\usepackage{multirow}
\usepackage{amssymb}
\usepackage[fleqn]{amsmath}
\usepackage{mathtools}
\usepackage{eqnarray}
\modulolinenumbers[5]

\usepackage{subfigure}
\usepackage{booktabs}
\usepackage{amssymb,mathtools,multirow}
\usepackage{amsmath,amssymb,amsfonts}
\usepackage{algorithmic}
\usepackage{algorithm}
\usepackage{xcolor}
\usepackage{algorithmic}
\usepackage{graphicx}
\usepackage{textcomp}
\usepackage{times}
\usepackage{soul}
\usepackage{url}
\usepackage[utf8]{inputenc}
\usepackage[small]{caption}
\usepackage{graphicx}
\usepackage{booktabs}
\usepackage{mathtools}
\usepackage{algorithm}
\usepackage{algorithmic}
\usepackage{makecell}
\usepackage{tabularx}
\usepackage{float}
\journal{}

\begin{document}

\begin{frontmatter}

\title{Assessor-Guided Learning for Continual Environments}

\author[Anwar]{Muhammad Anwar Ma'sum \fnref{fn1}}
\author[Mahardhika]{Mahardhika Pratama\fnref{fn1}}
\author[Edwin]{Edwin Lughofer}
\author[Weiping]{Weiping Ding}
\author[Wisnu]{Wisnu Jatmiko}

\fntext[fn1]{Authors share equal contributions.}
\fntext[fn2]{The source codes and raw numerical results can be downloaded from \url{https://github.com/anwarmaxsum/AGLA}}

\address[Anwar]{STEM, University of South Australia, Adelaide Australia. e-mail: muhammad\_anwar.masum@mymail.unisa.edu.au}
\address[Mahardhika]{STEM, University of South Australia, Adelaide Australia. e-mail: dhika.pratama@unisa.edu.au}
\address[Edwin]{the Institute of Mathematical Methods in Medicine and Databased Modeling, Johannes Kepler University, Linz, Austria. e-mail: edwin.lughofer@jku.at}
\address[Weiping]{Nantong University, China. e-mail: dwp9988@163.com}
\address[Wisnu]{Faculty of Computer Science, University of Indonesia, Jakarta, Indonesia. e-mail: wisnuj@cs.ui.ac.id}

\begin{abstract}
This paper proposes an assessor-guided learning strategy for continual learning where an assessor guides the learning process of a base learner by controlling the direction and pace of the learning process thus allowing an efficient learning of new environments while protecting against the catastrophic interference problem. The assessor is trained in a meta-learning manner with a meta-objective to boost the learning process of the base learner. It performs a soft-weighting mechanism of every sample accepting positive samples while rejecting negative samples. The training objective of a base learner is to minimize a meta-weighted combination of the cross entropy loss function, the dark experience replay (DER) loss function and the knowledge distillation loss function whose interactions are controlled in such a way to attain an improved performance. A compensated over-sampling (COS) strategy is developed to overcome the class imbalanced problem of the episodic memory due to limited memory budgets. Our approach, Assessor-Guided Learning Approach (AGLA), has been evaluated in the class-incremental and task-incremental learning problems. AGLA achieves improved performances compared to its competitors while the theoretical analysis of the COS strategy is offered. Source codes of AGLA, baseline algorithms and experimental logs are shared publicly in \url{https://github.com/anwarmaxsum/AGLA} for further study.

\end{abstract}

\begin{keyword}
Continual Learning, Meta-Learning, Lifelong Learning, Incremental Learning
\end{keyword}

\end{frontmatter}


\begin{table}[h]
\centering
\small
\begin{tabular}{ll}
\hline
\textbf{Abreviation} & \textbf{Description}                              \\
\hline
Task IL              & Task Incremental Learning                         \\
Class IL             & Class Incremental Learning                        \\
AGLA                 & Assessor-Guided Learning Approach                 \\
DER                  & Dark Experience Replay                            \\
ICARL                & Incremental Clasifier and Representation Learning \\
EEIL                 & End-to-End Incremental Learning                   \\
BIC                  & Bias Correction                                   \\
EWC                  & Elastic Weight Consolidation                      \\
SI                   & Synapse Intelligence                              \\
MAS                  & Memory Aware Synapse                              \\
LWF                  & Learning Without ForGetting                       \\
DMC                  & Deep Model Consolidation                          \\
HAL                  & Hindsight Anchor Learning                         \\
FIM                  & Fisher Information Matrix                         \\
SGD                  & Stochastic Gradient Descent                       \\
NAS                  & Neural architecture search                        \\
RLN                  & Representation Learning Network                   \\
PLN                  & Prediction Learning Network                       \\
COS                  & Compensated Over-Sampling                         \\
MLP                  & Multi-Layer Perceptron                            \\
CNN                  & Convolutional Neural Network                      \\
LSTM                 & Long-Short Term Memory                            \\
Kl divergence        & Kullback–Leibler divergence                      \\
MSE                  & Mean-Square Error                                 \\
Assr                 & Assessor                                          \\
Aug                  & Augmentation                                      \\
R.Tr                 & Random Transformation                            \\ \hline
\end{tabular}
\caption{Abbreviations list}
\end{table}

\section{Introduction}
\label{sec:introduction}

Continual learning problem aims to build a learning model throughout the life span of the model in use and gain its improved intelligence as the increase of learning tasks. Unlike conventional learning algorithms limited to only a single task, a continual learner is exposed to streaming tasks where each task possesses varying characteristics, i.e., different data distributions, different target classes, or combinations between distributional changes and class changes. A continual learner has to adapt quickly to new environments without losing its relevance to old tasks. This problem is not trivial for a deep neural network because of the catastrophic interference problem where old parameters are overwritten when learning a new task thereby losing its generalization power to the old tasks. Because of uncertain and possibly infinite problem sizes, a retraining process from scratch is undesired. In other words, the learning process occurs without complete accesses of old data samples. 

The continual learning problem goes one step toward human-like intelligence where the continual learner must be capable of accumulating knowledge from already seen experiences. As a result, this area has picked up substantial research attention. Existing works are categorized into three groups, regularization-based approach \cite{Kirkpatrick2017OvercomingCF}, structure-based approach \cite{Rusu2016ProgressiveNN} and memory-based approach \cite{LopezPaz2017GradientEM}. The regularization-based approach introduces an extra regularization term preventing important parameters of old tasks from deviations. This approach is simple to implement and computationally light. These approaches, however, do not scale well for large-scale problems because an overlapping region of all tasks is difficult to find with the regularization-based approach. The structure-based approach increases network capacity to deal with new tasks while isolating old network parameters to avoid the catastrophic forgetting problem. This approach is, however, computationally expensive and usually involves complex learning procedures. The memory-based approach takes another route where a small subset of old data samples are stored in the memory and interleaved with current samples when learning a new task. \emph{Conventional experience replay mechanism requires hundreds of samples to be stored in the memory thus incurring expensive memory footprints. There also exists the class imbalanced problem because old samples are often lower in quantity than new samples. The assessor-guided learning approach is put forward here to address these drawbacks where an assessor controls the learning process of the base learner via a soft-weighting mechanism of loss functions for every sample. Memory augmentation is applied in our method to address the class-imbalance problem between the current task and the previous tasks. The compensated over-sampling mechanism is integrated where self-corrections while over-sampling is performed to avoid the out-of-distribution cases.} 

Assessor-Guided Learning Approach (AGLA) is proposed here where a sequence-aware assessor is integrated to navigate the learning process of a base model to attain an improved learning performance. The assessor is trained with a meta-objective to boost the generalization power of the base model via a soft-weighting mechanism of loss functions for every sample. A high-quality sample is assigned with a high weight whereas a low weight is assigned to poor samples. High-quality samples are those leading to positive forward and backward transfers whereas poor samples are those imposing high losses associated with catastrophic forgetting. Two data subsets, training subset and validation subset, are created for each task to simulate the training-testing procedure \cite{Zheng2020DeepML} where both subsets represent current and old concepts. The concept of random transformation \cite{Volpi2021ContinualAO} is integrated to craft the validation subset. The training procedure follows the meta-learning principle where the training subset is used to train the base model in the outer loop while the assessor utilizes the validation subset for its updates in the inner loop. The assessor produces a set of weights, cross-entropy weight, dark experience replay (DER) weight and distillation weight controlling the interaction of loss functions and in turn the influence of every sample. This is made possible by formulating the loss function as a meta-weighted combination of the cross entropy loss function, the DER loss function \cite{Buzzega2020DarkEF} and the knowledge distillation loss function \cite{Rebuffi2017iCaRLIC}. The cross entropy loss focuses on current samples and past samples of the memory while the DER loss and the knowledge distillation loss targets past samples of the memory to maintain previous knowledge. In other words, the assessor steers a base model to address the stability- plasticity dillemma. It determines how much a base model should learn from the current condition and the past condition in respect to every sample. Note that both DER and knowledge distillation targets previous experiences because the number of previous tasks are larger than the current task but underrepresented in the training process, i.e., memory samples are much smaller than current samples. This aspect confirms the importance of meta-weighting mechanism regulating the influence of multi-objective functions seamlessly. 

The class imbalanced problem due to disproportionate proportions of memory samples and currents samples is tackled using the compensated over-sampling (COS) strategy where compensations are performed while over-sampling via well-known data augmentation protocols to protect against out-of-distribution augmented samples undermining model's generalization. A lemma analyzing such compensation w.r.t the bias-variance decomposition is provided as well as a theorem demonstrating reductions of MSEs as a result of the compensations is demonstrated.   

This paper conveys five major contributions:
\begin{enumerate}
  \item it puts forward the concept of assessor-guided continual learning where the sequence-aware assessor is deployed to guide the learning process of the base learner balancing the issue of stability and plasticity;
  \item the compensated over-sampling (COS) strategy is proposed to deal with the class imbalanced problem underpinned with theoretical analyses; 
  \item a meta-training strategy via a bi-level optimization is put forward to train the assessor where the concept of random transformation is implemented to craft the validation subset;
  \item a meta-weighted combination of three loss functions are proposed to train a base learner where each loss function is associated with either current or past conditions. This design offers an intuition of learning strategies for every sample, i.e, whether to focus on the current or previous contexts;
  \item the source codes of AGLA and other supporting data are made public in \url{ https://github.com/anwarmaxsum/AGLA} to enable further study.
\end{enumerate}

The advantage of AGLA over existing approaches has been numerically validated under the class-incremental and task-incremental learning configurations. It is demonstrated that AGLA delivers improvements compared to recently published algorithms in realm of average accuracy while each learning component contributes positively to the performance. AGLA achieves comparable average forgetting indexes where it mostly attains the second place and maintains decent performances with various memory sizes compared to baseline algorithms.

\section{Related Works}
\subsection{Continual Learning}
\noindent\textbf{Regularization-based method} designs a regularization term preventing important parameters of old tasks from deviations. Important parameters of old tasks are estimated and an important parameter matrix is integrated in the regularization term. Elastic Weight Consolidation (EWC) \cite{Kirkpatrick2017OvercomingCF} adopts the Fisher information matrix (FIM) to estimate the importance of network synapses. Synaptic intelligence (SI) \cite{Zenke2017ContinualLT} utilizes an accumulated gradient to quantify the significance of network parameters and incurs less expensive computation than FIM. An alternative is offered by memory aware synapses using an unsupervised and online approach \cite{Aljundi2018MemoryAS}. Learning without forgetting (LWF) \cite{Li2018LearningWF} applies the knowledge distillation approach making sure that current network outputs are close to previous network outputs. An online version of EWC is put forward in \cite{Schwarz2018ProgressC} where the parameter importance is approximated with the Laplace approximation. It is found that the regularization mechanism is better performed in the neuron level than the synaptic level because of the hierarchical nature of neural networks \cite{Paik2020OvercomingCF}. That is, the regularization step is achieved by controlling the learning rates of stochastic gradient descent (SGD) method. Similar approach is adopted in \cite{Mao2021ContinualLV} but it considers the common information of each task enabling a node to be shared by different tasks. This strategy is capable of scaling up the regularization-based method for large-scale problems. Another approach to scale the regularization-based approach is done with the classifier's projection \cite{Cha2021CPRCR}. This approach induces wide local optimum regions, i.e., overlapping region. The regularization-based approach depends on the task IDs and task boundaries, thus performing poorly for the class-incremental learning problems. 

\noindent\textbf{Structure-based approach} is pioneered by progressive neural networks \cite{Rusu2016ProgressiveNN} where new network components are added to deal with new tasks while freezing old network parameters to handle the catastrophic forgetting problem. The complexity of this approach grows as the increase of learning task. An error-based network growing method is put forward in \cite{Yoon2018LifelongLW}. It utilizes a selective-based retraining approach to handle the catastrophic forgetting problem. Learn-to-grow approach is developed in \cite{Li2019LearnTG} where neural architecture search (NAS) is integrated to obtain the best network configuration of new tasks while isolating old network parameters. The same concept is implemented for graph continual learning in \cite{Rakaraddi2022ReinforcedCL}. The Bayesian approach is incorporated in \cite{Xu2021AdaptivePC}. The idea is akin to \cite{Li2019LearnTG} but the Bayesian approach is adopted instead of NAS to find the best network structure. These approaches are computationally prohibitive and calls for the presence of tasks IDs and task's boundaries. \cite{Pratama2021UnsupervisedCL,Ashfahani2022UnsupervisedCL} offer a data-driven structural learning approach for unsupervised continual learning where the bias-variance decomposition is put forward to grow or prune the network structure. The catastrophic forgetting problem is dealt with the centroid-based experience replay mechanism \cite{Pratama2021UnsupervisedCL} or the knowledge-distillation approach \cite{Ashfahani2022UnsupervisedCL}. Although it is free of the task IDs and task boundaries for learning and predicting, the data-driven structural learning technique does not guarantee an optimal structure. 

\noindent\textbf{Memory-based approach} replays old samples stored in the memory when learning new tasks. iCaRL \cite{Rebuffi2017iCaRLIC} utilizes the exemplar set of each class where the classification step is performed via the nearest-mean strategy. GEM \cite{LopezPaz2017GradientEM} and A-GEM \cite{Chaudhry2019EfficientLL} store past examples to determine the forgetting case used to constrain the model update. HAL \cite{Chaudhry2021UsingHT} puts forward the concept of anchor points optimized to maximize the forgetting case via a bilevel optimization approach. Prediction should not change to these anchor points when learning new tasks. DER \cite{Buzzega2020DarkEF} combines the concept of knowledge distillation and experience replay. \cite{VinciusdeCarvalho2022ClassIncrementalLV} proposes the concept of knowledge amalgamation as a post-processing approach of class-incremental continual learning. These methods suffer from the class imbalanced problem since the size of memory buffer is much less than new samples. AGLA presents an extension of the memory-based approach where a sequence-aware assessor is deployed to guide the learning process of the base model. The assessor not only regulates the influence of every sample such that only positive samples are learned but also selects a suitable learning strategy of a given sample where the interaction of loss functions is controlled in a seamless manner. The class imbalanced issue is tackled with the COS method handling out of distributions of augmented samples.  


\subsection{Meta Learning}
Meta learning also known as learning-to-learn aims to learn an algorithm to improve the learning performance of another learning algorithm. This approach has been adopted in the continual learning problem in \cite{Javed2019MetaLearningRF,Pham2021ContextualTN,Dam2022ScalableAO}. \cite{Javed2019MetaLearningRF} creates two networks: representation learning network (RLN) and prediction learning network (PLN). RLN is trained with a meta-objective later used as representation of PLN. \cite{Pham2021ContextualTN} integrates the controller network generating scaling and shifting parameters to generate task-specific features. \cite{Dam2022ScalableAO} applies the meta-learning concept to scale adversarial continual learning \cite{Ebrahimi2020AdversarialCL} for online continual learning cases. \cite{Zheng2020DeepML} introduces the assessor-guided learning principle in a single-task metric learning problem to address the flaws of hard mining. Our approach distinguishes itself from aforementioned methods where the meta-learning concept is developed to construct an assessor controlling the stability and plasticity of a base network in multi-task continual learning problems.

\section{Problem Formulation}
Supervised continual learning problems are considered here where a model is trained to a sequence of fully labelled tasks $\mathcal{T}_1,\mathcal{T}_2,...,\mathcal{T}_{K}$. 
Each task $\mathcal{T}_{k}=\{x_n,y_n\}_{n=1}^{N_k},$ $k\in\{1,...,K\}$ is sampled from the i.i.d distribution $\mathcal{D}_k$ where $x_{n}\in\mathcal{X}$ stands for an input sample and $y_{n}\in\mathcal{Y}$ denotes its true class label. $K, N_k$ respectively denote the number of tasks which might be infinite and the size of $k-th$ task. Each task does not possess the same characteristic causing non-stationary environments to a continual learning model $g_{\phi}(f_{\theta}(.))$ with feature extractor's parameters $\theta$ and classifier's parameters $\phi$. There exist three continual learning variants in the literature: domain-incremental, task-incremental and class-incremental \cite{Ven2019ThreeSF}. The domain-incremental problem refers to different data distributions of each task $P(X,Y)_{k}\neq P(X,Y)_{k+1},k\in\{1,...,K\}$ while having the same problem structure, i.e., input and target variables. The task-incremental and class-incremental problems feature different class labels of each task. Suppose that $L_{k},L_{k'}, k,k'\in\{1,..,K\}$ stand for label sets of the $k-th$ task and the $k'-th$ task, $\forall k,k',L_k\cap L_{k'}=\emptyset$. 
The difference between the task-incremental and class-incremental problems lies in \textbf{the absence of the task IDs} for the class-incremental problem. A prediction relies on a single classifier $g_{\phi}(.)$ rather than one classifier per task $g_{\phi_{k}}(.)$. A task $\mathcal{T}_k$ is accessed at the $k-th$ session and discarded once completed. This issue leads to the catastrophic interference problem where learning a new task $\mathcal{T}_k$ overwrites previously valid parameters.

\section{Learning Procedure of AGLA}
AGLA is developed from the assessor-guided learning method where an assessor $\kappa_{\psi}(.)$ is deployed to guide the learning process of the base learner $g_{\phi}(f_{\theta}(.))$. The assessor produces a set of weights, the cross entropy weight, the DER weight, and the distillation weight, for every sample. The soft-weighting mechanism not only controls the pace and direction of the learning process based on the quality of a data sample, but also governs the interaction of each loss function making possible for an adaptive selection of a suitable learning strategy for every sample, i.e., which losses to be favoured. The compensated over-sampling (COS) strategy is integrated to cope with the class-imbalanced issue on continual learning where corrections are performed while over-sampling to prevent the adverse impacts of out-of-distribution samples.  
\subsection{Loss Function}
The cost function is formulated as a meta-weighted combination of the cross entropy loss function, the DER loss function \cite{Buzzega2020DarkEF} and the knowledge distillation loss function \cite{Rebuffi2017iCaRLIC}. 
The three loss functions have been rigorously validated where the absence of one term results in significant performance drops as shown in our ablation study.
\begin{equation}\label{loss}
\begin{split}
    \mathcal{L}_{bl}=\underbrace{\mathbb{E}_{(x,y)\backsim \mathcal{D}_k\cup\mathcal{\hat{M}}_{k-1}}[\alpha l(s_{W}(o),y)]}_{L_{CE}}+\underbrace{\mathbb{E}_{(x,y)\backsim\mathcal{\hat{M}}_{k-1}}\lambda[||o-h||+l(s_{W}(o),y)]}_{L_{DER++}}+\\\underbrace{\mathbb{E}_{(x,y)\backsim\mathcal{\hat{M}}_{k-1}}[\pi l(o,h)]}_{L_{distill}}  
\end{split}
\end{equation}
\begin{align*}\label{factor}
    \lambda = \begin{cases} 0, & k=1\\
    k*\beta, & k\geq 1
    \end{cases}, && \pi=\begin{cases} 0, &k=1\\
    k*\gamma, & k\geq 1
    \end{cases}
\end{align*}
where $\alpha,\lambda,\pi$ are respectively the cross entropy weight, the DER weight and the distillation weight generated by the sequence-aware assessor $\{\alpha,\beta,\gamma\}=\kappa_{\psi}(x)\in[0,1]$. $o=g_{\phi}(f_{\theta}(.))$ is a presoftmax response known as output logit. 
In addition to determine the sample's influence in the learning process, these three weights determine the learning strategies for a sample of interest, i.e., it is capable of selecting a suitable loss function for every sample. Note that the soft-weighting mechanism is applied and provides better flexibility than the binary hard-sampling mechanism because the learning process of the base learner is governed in a smooth manner.  $h=g_{\phi}(f_{\theta}(x))_{k-1}$ and $||.||_2$ is the L-2 distance function. Note that the L-2 distance function functions similarly as the KL divergence. No softmax function is applied to avoid the effect of squashing function and the real labels are included here $L_{DER++}$ to prevent the distribution shift \cite{Buzzega2020DarkEF}. Both the DER loss $L_{DER++}$ and the distillation loss $L_{distill}$ aim to maintain the stability of previously learned knowledge while the cross entropy loss $L_{CE}$ aims to enhance the plasticity to new knowledge. The three losses are meta-weighted by the assessor to achieve a proper tradeoff of the plasticity and the stability. Two loss functions, focusing on old tasks, the DER loss function and the distillation loss function, are integrated here because the number of old tasks are usually larger than the current task but under-represented in the learning process due to a small memory size. In addition, $L_{CE}$ involves both current and old samples to avoid biased responses toward new classes and has been a common design choice in the literature \cite{Wu2019LargeSI,Castro2018EndtoEndIL}. The use of meta-weights also avoids a tweaking problem in continual learning algorithms using a large weight $>100$ to the losses corresponding to the past states. 

The base learner $g_{\phi}(f_{\theta}(.))$ is formulated as the MLP or CNN where the feature extractor $f_{\theta}(.)$ is created as stacked linear layers or convolutional layers while the classifier $g_{\phi}(.)$ is formed as a fully connected layer. One classifier per task $g_{\phi_k}(.)$ is used for the task-incremental learning problem while a single classifier $g_{\phi}(.)$ is applied for the class-incremental learning problems without any task IDs. The assessor $\kappa_{\psi}(.)$ is formed as LSTM followed by a fully connected layer with the sigmoid activation function at the last layer. The convolutional layers are integrated as the feature extractor in the assessor. The use of LSTM aims to provide short-term memory, hidden state, and long-term memory, cell state, making possible for a sequence of past weights to be preserved in the memory. $l(.)$ is a cross-entropy loss function and $s_{W}(.)$ is a softmax layer parameterized by $W$. 

\subsection{Meta-Training Strategy}
The assessor is trained with a meta-objective to boost the learning process of the base learner, i.e., the influence of a current sample and an old sample as well as the interaction of the three loss functions are controlled to achieve a tradeoff between the issue of plasticity and stability. That is, the assessor $\{\alpha,\beta,\gamma\}=\kappa_{\psi}(x)\in[0,1]$ controls the speed, direction and learning strategies of the base learner. Suppose the current $k-th$ task, two data partitions, training set and validation set, are created where the training set is constructed from current samples and memory samples $\mathcal{T}_{train}^{k}=\mathcal{T}_{train}^k\cup\mathcal{\hat{M}}_{k-1}$ whereas the validation set is created by applying random transformation to the training set $\mathcal{T}_{val}^{k}=\{T(x_i),y_i\}_{i=1}^{N_{train}^{k}}\backsim\mathcal{T}_{train}^{k}$. $T(.)$ is the random transformation operator such as color/geometric transformation or noise injection \cite{Volpi2021ContinualAO} and $N_{train}^{k}$ is the size of the training set. We follow the same way as \cite{Volpi2021ContinualAO} where $T(.)$ is taken from one of the possible transformation sets $\Phi$, i.e., three transformations, image invert, Gaussian noise perturbation, RGB-rand perturbation are applied here to the original training samples. $\mathcal{\hat{M}}_{k-1}$ denotes an augmented memory set including those of the data augmentation procedure. The training set $\mathcal{T}_{train}^{k}$ is exploited to update the base learner whereas the validation set $\mathcal{T}_{val}^{k}$ is used to train the assessor. 

The training process of the base learner and the assessor is formulated as a bi-level optimization problem \cite{Pham2021ContextualTN} where the base learner is updated using the training set while the assessor is trained using the validation set. That is, the bi-level optimization problem is formulated as follows:
\begin{equation}\label{bilevel}
    \begin{split}
        \min_{\psi}\mathbb{E}_{(x,y)\backsim\mathcal{T}_{val}^{k}}[\mathcal{L}_{bl}(g_{\phi^*}(f_{\theta^*}(x)),y)]\\
        s.t,\phi^*,\theta^*=\arg\min_{\phi,\theta}\mathbb{E}_{(x,y)\backsim\mathcal{T}_{train}^{k}}[\mathcal{L}_{bl}(g_{\phi}(f_{\theta}(x)),y)]
    \end{split}
\end{equation}
where $\{\phi^*,\theta^*\}$ stand for optimal base parameters with respect to the current assessor $\psi$. The meta-learning strategy is implemented here because of the absence of ground truth of the assessor, i.e., the ideal weights. This case implies the optimal parameters of the assessor $\psi^*$ minimizing the validation loss of the base network. The bi-level optimization approach is solved by first updating the assessor. That is, the base learner is evaluated with the validation set $\mathcal{T}_{val}^k$ returning the validation loss. The validation loss is used to update the assessor. In other words, the learning process of the assessor aims to minimize a meta-objective as follows:
\begin{equation}\label{loss_ass} \psi^*=\arg\min_{\psi}\sum_{\mathclap{(x,y)\in\mathcal{T}_{val}^{k}}}\mathcal{L}_{bl}(g_{\phi}(f_{\theta}(x)),y)
\end{equation}
The validation loss is differentiable with respect to the assessor because the validation loss is a factor of the three weights $\alpha,\beta,\gamma$ produced by the assessor. Since the optimal base model depends on the assessor, the assessor has to be updated first:
\begin{equation}\label{inner}
  \psi'=\psi-\eta\sum_{\mathclap{(x,y)\in\mathcal{T}_{val}^k}}\nabla_{\psi}\mathcal{L}_{bl}(g_{\phi}(f_{\theta}(x)),y)  
\end{equation}
where $\eta$ is a learning rate. 
Once completing the assessor update, the base learner is adjusted using the meta-weights generated by the updated assessor:
\begin{equation}\label{outer}
        \{\phi,\theta\}=\{\phi,\theta\}-\mu\sum_{\mathclap{(x,y)\in\mathcal{T}_{train}^{k}}}\nabla_{\{\phi,\theta\}}[\mathcal{L}_{bl}(g_{\phi}(f_{\theta}(x)),y)]
\end{equation}
where $\mu$ is a learning rate. 
This strategy implies a joint update of the base learner and the assessor where the adjustment of base learner involves the tuning step of the assessor. It guarantees an improved performance of the base learner because it is undertaken using the cross entropy weight $\alpha$, the DER weight $\beta$, and the distillation weight $\gamma$ produced by an updated assessor $\{\alpha,\beta,\gamma\}=\kappa_{\psi'}(x)$. This strategy also overcomes the tweaking issue of the memory-based approach in continual learning where memory samples are assigned with large weights, e.g., $>100$ confirming over-dependence on the experience replay strategy while ignoring new tasks.
\subsection{Compensated Over-Sampling Approach}
The main challenge of the memory-based approach is to resolve the class imbalanced problem because only a small fraction of old samples are stored. The class imbalanced problem results in a bias problem where a model favours new classes because old samples of the memory are under-represented in the data batches. \cite{Wu2019LargeSI} shows that the last fully connected layer is heavily biased because it is not shared during the training process. We apply the over-sampling approach to correct the bias problem. That is, a memory set $\mathcal{M}_{k-1}$ is augmented using the random transformation method, thus generating an augmented memory set $\mathcal{\hat{M}}_{k-1}$. The augmented memory set rectifies disproportionate class proportions between new classes and old classes. 

A naive oversampling approach might induce out-of-distribution samples undermining model's generalization power \cite{Wang2021ImprovingSL}. This paper proposes a compensated over-sampling (COS) strategy for continual learning where the learning process of augmented samples is compensated to offset out-of-distribution samples. Ideally, the over-sampling approach via a family of the transformation functions $T(.)$ should sample from the same distribution $p^*$ and semantic of an original samples $x_i$. Because such distribution $p^*$ is unknown in practise, we rely on a good assumption of heuristically chosen $T(.)\backsim \tilde{p}$:
\begin{equation}
    \mathcal{L}^*=\sum_{i=1}^{|\mathcal{M}_{k-1}|}\sum_{k=1}^{M}\mathcal{L}_{bl}(T(x_i)_{k},y_i;\theta,\phi)\tilde{p}_{k}\frac{p^*_{k}}{\tilde{p}_k}
\end{equation}
The ratio $p_k^*/\tilde{p}_k$ is hard to access in practise. Alternatively, we define $w_k=\frac{p_k^*}{\tilde{p}_k}$ following the Radon Nikodym derivative \cite{Wang2021ImprovingSL} and analyze how it affects the MSE of the deep model. The following lemma describes the MSE of the deep estimator. 

\noindent\textbf{Lemma 1}\cite{Wang2021ImprovingSL}: Define $\theta_0=\arg\min_{\theta}\mathbb{E}_x\mathcal{L}_{\theta}(x), \theta_{G}=\arg\min_{\theta}\mathbb{E}_x\int\mathcal{L}_{\theta}(x,n)dp^*$, and $\hat{\theta}_{G}=\arg\min_{\theta}\mathcal{L}^*$. Let $V_0$ be the Hessian of $\theta\rightarrow\mathbb{E}_x\mathbb{L}_{\theta}(x)$ at $\theta_0$, and $V_G$ the Hessian of $\theta\rightarrow\mathbb{E}_x[\int\mathcal{L}_{\theta}(x,n)d\tilde{p}(n)w(n)]$ at $\theta_G$. Let $M_0(x)=\nabla\mathcal{L}_{\theta_0}(x)\nabla\mathcal{L}_{\theta_0}(x)^{T}$ and $M_G(x)=\nabla\mathcal{L}_{\theta_G}(x)\nabla\mathcal{L}_{\theta_G}(x)^{T}$ where $\nabla\mathcal{L}_{\theta_0}(x)$ and $\nabla\mathcal{L}_{\theta_G}(x)$ correspond to the gradients of $\mathcal{L}_{\theta}(x)$ at $\theta=\theta_0$ and $\theta=\theta_G$ respectively. Suppose $M_G(x,n)=\nabla\mathcal{L}_{{\theta}_{G}}(x,n)\nabla\mathcal{L}_{{\theta}_{G}}(x,n)^{T}$. $tr(X)$ stands for the trace of matrix $X$. Hence, with $C$ a constant invariant of $w$, under mild conditions, we obtain:

\begin{equation}
\begin{split}
    MSE(\hat{\theta}_{G})\backsim C + ||\theta_G - \theta_0||_2^2+\frac{1}{N}\mathbb{E}_x[\int tr(V_G^{-1}(M_G(x,n)\\-M_G(x))V_G^{-1})d\tilde{p}(n)w(n)]\\+\frac{1}{N}\mathbb{E}_x[tr(V_G^{-1}(M_G(x)-M_0(x))V_G^{-1})]
\end{split}
\end{equation}
\begin{equation}
    \begin{split}
        +\frac{1}{N}tr((V_G^{-1}-V_{0}^{-1})Cov_{x}\nabla\mathcal{L}_{{\theta}_{0}}(x)(V_G^{-1}-V_{0}^{-1}))
    \end{split}
\end{equation}
\begin{equation}
    \begin{split}
        -\frac{1}{N}tr(V_{G}^{-1}\mathbb{E}_x[Cov_{w}\nabla\mathcal{L}_{{\theta}_{G}}(x,n)]V_{G}^{-1})
    \end{split}
\end{equation}
where $Cov_{w}(\nabla\mathcal{L}_{{\theta}_{G}}(x,n))$ is the covariance matrix of $\mathcal{L}_{\theta}(x,n)$ at $\theta_{G}$ under measure $p^*(n)$. 

This lemma tells that large data variance via sampling from $p^*$ is desired to decrease the model's variance. Another case also explains the increase of MSEs or biases if there exist significant differences between the gradient of augmented samples and those of original samples. The key observation is the presence of $w(n)$ which can be benefited to control the bias-variance trade-off. In other words, it is capable of compensating possible out of distribution augmented samples. We apply the same principle as \cite{Wang2021ImprovingSL} where $w_k$ is determined as the likelihood ratio in sampling $T(x_i)_k$ to reduce the MSE in Lemma 1. Suppose that $z_{i,k}=f_{\theta}(T(x_i)_k)$, $w_{i,k}$ is arranged

\begin{equation}\label{ratio}
\begin{split}
    \frac{p^*}{\tilde{p}}\propto w_{i,k}=\exp{-(z_{i,k}-\mu_i)(\tau\Sigma)^{-1}(z_{i,k}-\mu_i)}
\end{split}
\end{equation}
\begin{equation}
    \begin{split}
        \mu_i=\frac{1}{M}\sum_{k=1}^{M}z_{i,k};\Sigma=\frac{1}{NM}\sum_{i}^{N}\sum_{k}^{M}(z_{i,k}-\mu_i)(z_{i,k}-\mu_i)^{T}
    \end{split}
\end{equation}
where $k\in\{1,...,M\}$, $M$ denotes the number of random transformations applied an image $x_i$ and $\tau$ is a temperature that regulates the influence of the distance measure. $w_{i,k}$ is further normalized across a data batch leading to $\overline{x}_i$:
\begin{equation}\label{normalized_ratio}
  \frac{p^*}{\tilde{p}}\propto \overline{w}_{i,k}=\frac{w_{i,k}}{\sum_{i=1}^{N}\sum_{k=1}^{M}w_{i,k}}  
\end{equation}
The normalized ratio is applied to govern the learning process of augmented samples:
\begin{equation}\label{weighted_loss}
    \mathcal{L}_{bl}=\sum_{i=1}^{N}\sum_{k=1}^{M}\overline{w}_{i,k}\mathcal{L}_{bl}
\end{equation}
At first, a forward propagation is performed to produce $\mu_{i},\Sigma$ leading to $\overline{w}_{i,k}$. $\overline{w}_{i,k}$ is frozen afterward and inserted during the learning process of $\mathcal{L}_{bl}$. A theorem can be derived. 

\noindent\textbf{Theorem 1} \cite{Wang2021ImprovingSL}: given mild conditions, the learning process of augmented samples via \eqref{weighted_loss} where $\overline{w}_{i,k}$ is determined via \eqref{normalized_ratio}, \eqref{ratio} leads to reductions of MSEs as per Lemma 1. 

Proofs of Theorem 1 and Lemma 1 are provided in the supplemental. Algorithm 1 exhibits the pseudo-code of AGLA.

\begin{algorithm}
\caption{Learning Policy of AGLA}
\begin{algorithmic}
\STATE\textbf{Input}: continual dataset $\mathcal{D}$, learning rates $\mu,\eta$, size of pseudo memory $m$, iteration number $E$ 
\STATE\textbf{Output}: parameters of base learner $\{\phi,\theta\}$, parameters of the assessor $\psi$
\FOR{$k=1$ to $K$}
\STATE $\mathcal{\hat{M}}_{k-1}=Augment(\mathcal{M}_{k-1})$ /* Augment the memory to address the class imbalanced problem/*
\STATE$\mathcal{T}_{train}^{k}=\mathcal{\hat{M}}_{k-1}\cup\mathcal{T}_{k}$ /*Construct the training set/*
\STATE$\mathcal{T}_{val}^{k}=\{T(x_i),y_i\}_{i=1}^{N_{train}^{k}}\backsim\mathcal{T}_{train}^{k}$ /*Apply random transformation to construct the validation set/*
\FOR{$e=1$ to $E$}
\STATE Update assessor parameters $\psi$ using \eqref{inner}
\STATE Update base learner parameters $\{\phi,\theta\}$ using \eqref{outer}
\ENDFOR
\STATE $B_k=Sample(\mathcal{T}_k)$ /* Perform reservoir sampling on the current task /*
\STATE $\mathcal{M}_k=\mathcal{M}_{k-1}\cup B_k$ /*Update the current memory/*
\ENDFOR
\end{algorithmic}
\end{algorithm}

\begin{figure}
\includegraphics[width=1.0\textwidth]{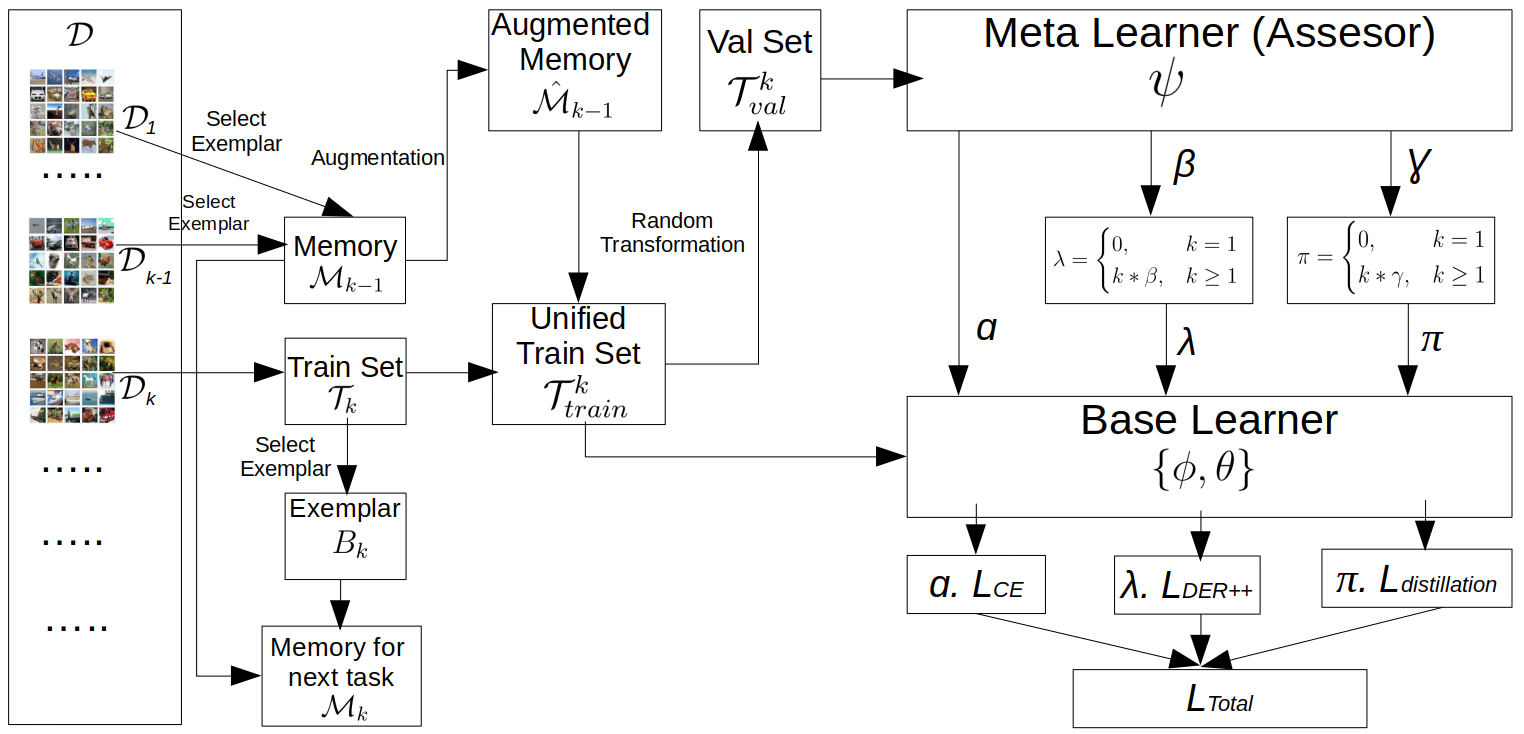}
\caption{The flow-chart of the proposed method (AGLA).}
\label{Fig:AGLA diagram}
\end{figure}

\begin{table}
\begin{center}
\begin{tabular}{||c c c c||} 
 \hline
 Problem & nTasks & nClasses / Task & CL Setting \\ [0.5ex] 
 \hline
 SMNIST & 5 & 2 & Class and Task - IL \\ 
 \hline
 SCIFAR10 & 5 & 2 & Class and Task - IL \\
 \hline
 SCIFAR100 & 10 & 10 & Class and Task - IL \\
 \hline
 SMINIIMAGENET & 10 & 10 & Class and Task - IL \\
 \hline
\end{tabular}
\caption{Details of Continual Learning Problems, nTasks: number of tasks, nClasses / Task: number of classes per task, CL Setting: Continual Learning Setting}
\label{tab:problems}
\end{center}
\end{table}

\begin{table}
\centering
\footnotesize
\begin{tabularx}{1.0\textwidth} { 
   >{\raggedright\arraybackslash}p{0.4\textwidth}X 
   >{\raggedright\arraybackslash}X 
   >{\raggedright\arraybackslash}X 
}
\hline
\textbf{Methods}                                                                                                                         & \textbf{Params}            & \textbf{Value} \\ \hline
\multirow{24}{4em}{\begin{tabular}[c]{@{}l@{}}All: AGLA, DER, \\ BIC, ICARL, EEIL, \\ EWC, LWF, MAS,\\  SI, Fnetuning, Joint\end{tabular}} & batch$\_$size                & 100            \\ \cline{2-3}  & clipping                   & 10000          \\ \cline{2-3}  & eval$\_$on$\_$train            & False          \\ \cline{2-3}  & fix$\_$bn                    & False          \\ \cline{2-3}  & gridsearch$\_$tasks          & -1             \\ \cline{2-3}  & keep\_existing$\_$head       & False          \\ \cline{2-3}  & last$\_$layer$\_$analysis      & False          \\ \cline{2-3}  & log                        & {[}'disk'{]}   \\ \cline{2-3}  & lr                         & 0.05           \\ \cline{2-3}  & lr$\_$factor                 & 1              \\ \cline{2-3}  & lr$\_$min                    & 0.0001         \\ \cline{2-3}  & lr$\_$patience               & 5              \\ \cline{2-3}  & momentum                   & 0.9            \\ \cline{2-3}  & multi$\_$softmax             & False          \\ \cline{2-3}  & nepochs                    & 100            \\ \cline{2-3}  & network                    & resnet18       \\ \cline{2-3}  & no$\_$cudnn$\_$deterministic   & False          \\ \cline{2-3}  & num$\_$workers               & 4              \\ \cline{2-3}  & pin$\_$memory                & False          \\ \cline{2-3}  & pretrained                 & False          \\ \cline{2-3}  & use$\_$valid$\_$only           & False          \\ \cline{2-3}  & warmup$\_$lr$\_$factor         & 1              \\ \cline{2-3}  & warmup$\_$nepochs            & 0              \\ \cline{2-3}  & weight$\_$decay              & 0.0002         \\ \hline
\multirow{2}{4em}{\begin{tabular}[c]{@{}l@{}}Memory Based: AGLA, DER \\ BIC, ICARL, EEIL\end{tabular}}                                 & num$\_$exemplars$\_$per$\_$class & 50             \\ \cline{2-3}  & exemplar$\_$selection        & random         \\ \hline
\multirow{4}{4em}{BIC}                                                                                                                     & T                          & 2              \\ \cline{2-3}  & lamb                       & -1             \\ \cline{2-3}  & num$\_$bias$\_$epochs          & 200            \\ \cline{2-3}  & val$\_$exemplar$\_$percentage  & 0.1            \\ \hline
{ICARL}                                                                                                                                    & lamb                       & 1              \\ \hline
\multirow{5}{4em}{EEIL}                                                                                                                    & T                          & 2              \\ \cline{2-3}  & lamb                       & 1              \\ \cline{2-3}  & lr$\_$finetuning$\_$factor     & 0.01           \\ \cline{2-3}  & nepochs$\_$finetuning        & 40             \\ \cline{2-3}  & noise$\_$grad                & FALSE          \\ \hline
\multirow{4}{4em}{EWC}                                                                                                                     & alpha                      & 0.5            \\ \cline{2-3}  & fi$\_$num$\_$samples           & -1             \\ \cline{2-3}  & fi$\_$sampling$\_$type         & max$\_$pred      \\ \cline{2-3}  & lamb                       & 5000           \\ \hline
\multirow{2}{4em}{LWF}                                                                                                                     & T                          & 2              \\ \cline{2-3}  & lamb                       & 1              \\ \hline
\multirow{3}{4em}{MAS}                                                                                                                     & alpha                      & 0.5            \\ \cline{2-3}  & fi$\_$num$\_$samples           & -1             \\ \cline{2-3}  & lamb                       & 1              \\ \hline
\multirow{2}{4em}{SI}                                                                                                                      & damping                    & 0.1            \\ \cline{2-3}  & lamb                       & 0.1            \\ \hline
\end{tabularx}
\label{tab:hyperparameters}
\caption{Hyper-parameters of Consolidated Algorithms}
\end{table}

\begin{table*}[t]
\resizebox{\textwidth}{!}{
\begin{tabular}{lcccccccc}
\hline
{} & \multicolumn{2}{c}{{\textbf{S-CIFAR-100}}}                         & \multicolumn{2}{c}{{\textbf{S-CIFAR-10}}}                        & \multicolumn{2}{c}{{\textbf{S-MINI-IMAGENET}}}                    & \multicolumn{2}{c}{{\textbf{S-MNIST}}}                             \\ \cline{2-9} 
{{\textbf{Method}}} & {Task IL}             & {Class IL}            & {Task IL}            & {Class IL}           & {Task IL}             & {Class IL}           & {Task IL}             & {Class IL}            \\ \hline
{Finetuning}                        & {49.01$\pm$1.84}          & {24.21$\pm$0.76}          & {74.18$\pm$5.18}         & {42.88$\pm$1.57}         & {44.06$\pm$1.78}          & {20.99$\pm$0.89}         & {78.33$\pm$4.28}          & {40.92$\pm$3.24}          \\ 
{Joint}                             & {75.45$\pm$1.71}          & {55.88$\pm$2.03}          & {95.86$\pm$1.53}         & {86.94$\pm$0.93}         & {66.45$\pm$1.95}          & {44.43$\pm$1.55}         & {99.88$\pm$0.06}$\ast$              & {99.62$\pm$0.09}$\ast$              \\ \hline
{EWC\cite{Kirkpatrick2017OvercomingCF}}                               & {53.46$\pm$2.02}          & {26.18$\pm$1.21}          & {77.16$\pm$6.22}         & {40.51$\pm$4.54}         & {48.41$\pm$2.36}          & {23.26$\pm$0.94}         & {83.07$\pm$2.83}          & {43.19$\pm$3.57}         \\ 
{LWF\cite{Li2018LearningWF}}                               & {51.28$\pm$14.82}         & {26.61$\pm$4.73}          & {91.14$\pm$3.61}         & {56.32$\pm$1.89}         & {52.04$\pm$2.5}           & {24.34$\pm$1.33}         & {79.84$\pm$12.60}         & {51.38$\pm$8.69}           \\ 
{MAS\cite{Aljundi2018MemoryAS}}                               & {52.3$\pm$1.95}           & {26.01$\pm$1.15}          & {76.72$\pm$5.21}         & {38.94$\pm$4.77}         & {46.57$\pm$2.36}          & {22.16$\pm$1.08}         & {86.50$\pm$4.83}          & {44.97$\pm$2.31}          \\ 
{SI\cite{Zenke2017ContinualLT}}                                & {52.19$\pm$1.85}          & {25.56$\pm$1}             & {76.44$\pm$5.67}         & {39.2$\pm$2.6}           & {46.72$\pm$2.22}          & {22.15$\pm$1.06}         & {82.10$\pm$3.13}          & {41.48$\pm$1.20}             \\ 
 {DMC\cite{zhang2020class}}                               &  {62.31$\pm$1.91}          &  {35.72$\pm$2.07}          & {81.90$\pm$11.40}          & {52.70$\pm$10.10}          & {53.42$\pm$1.61}         & {27.91$\pm$1.73}          & {77.90$\pm$15.50}          &  {48.50$\pm$15.30}           \\ \hline
{EEIL\cite{Castro2018EndtoEndIL}}                              & {67.24$\pm$1.56}          & {40.75$\pm$1.49}          & {93.99$\pm$1.67}         & {71.2$\pm$2.93}          & {57.99$\pm$2.51}          & {30.84$\pm$1.69}         & {86.51$\pm$6.58}           & {72.74$\pm$11.74}         \\ 
{ICARL\cite{Rebuffi2017iCaRLIC}}                             & {67.13$\pm$1.51}          & {44.15$\pm$2.02}          & {93.91$\pm$2.35}         & {78.54$\pm$1.4}          & {58.53$\pm$2.63}          & {34.93$\pm$2.14}         & {99.79$\pm$0.07}          & {98.67$\pm$0.15}          \\ 
{BIC\cite{Wu2019LargeSI}}                               & {67.82$\pm$2.89}          & {44.63$\pm$3.02}          & {89.41$\pm$10.59}        & {71.1$\pm$15.41}         & {59.16$\pm$3.05}          & {35.19$\pm$2.11}         & {76.48$\pm$4.52}          & {51.02$\pm$2.25}          \\ 
{DER++\cite{Buzzega2020DarkEF}}                               & {65.89$\pm$1.84}          & {41$\pm$1.72}             & {94.7$\pm$1.6}           & {77.42$\pm$1.98}         & {55.91$\pm$2.12}          & {30.78$\pm$1.3}          & {99.77$\pm$0.08}          & {98.54$\pm$0.28}           \\ 
 {HAL\cite{chaudhry2021using}}                               & {53.77$\pm$10.76}          & {30.08$\pm$8.52}          & {74.60$\pm$3.64}          & {46.90$\pm$1.57}          & {45.94$\pm$8.11}         & {22.69$\pm$5.06}          & {71.30$\pm$4.55}          & {45.80$\pm$0.19}           \\
{\textbf{AGLA (Ours)}}                     & {\textbf{69.51$\pm$1.68}} & {\textbf{46.95$\pm$2.19}} & {\textbf{95.64$\pm$1.53}} & {\textbf{82.64$\pm$1.14}} & {\textbf{60.54$\pm$2.09}} & {\textbf{36.23$\pm$1.66}} & {\textbf{99.86$\pm$0.04}} & {\textbf{98.86$\pm$0.18}} \\ \hline
\end{tabular}
}

\caption{Classification result (accuracy $\%$) for continual learning benchmarks, averaged across 5 runs. $\ast$ 4 tasks only due to crash}
\label{tab:average_accuracy_all}
\end{table*}

\begin{table*}[t]
\resizebox{\textwidth}{!}{
\begin{tabular}{lcccccccc}
\hline
{}                                  & \multicolumn{2}{c}{{\textbf{S-CIFAR-100}}}                        & \multicolumn{2}{c}{{\textbf{S-CIFAR-10}}}                         & \multicolumn{2}{c}{{\textbf{S-MINI-IMAGENET}}}                    & \multicolumn{2}{c}{{\textbf{S-MNIST}}}                           \\ \cline{2-9} 
{{\textbf{Method}}} & {Task IL}            & {Class IL}            & {Task IL}            & {Class IL}            & {Task IL}            & {Class IL}            & {Task IL}            & {Class IL}           \\ \hline
{Finetuning}                        & {31.28$\pm$1.72}         & {59.54$\pm$2.14}          & {37.71$\pm$9.5}          & {77.9$\pm$9.12}           & {24.58$\pm$1.37}         & {50.11$\pm$2.41}          & {37.67$\pm$7.78}         & {69.24$\pm$8.44}         \\ 
{Joint}                             & {-1.21$\pm$0.2}          & {5.79$\pm$0.59}           & {-0.73$\pm$0.17}         & {6.26$\pm$1.63}           & {-1.14$\pm$0.17}         & {6.23$\pm$1.18}           & {-0.02$\pm$0.02}$\ast$                & {0.15$\pm$0.06}$\ast$                \\ \hline
{EWC\cite{Kirkpatrick2017OvercomingCF}}                               & {19.23$\pm$2.37}         & {39.62$\pm$2.9}           & {29.47$\pm$11.44}        & {50.86$\pm$13.93}         & {14.6$\pm$1.56}          & {32.51$\pm$2.8}           & {29.66$\pm$4.69}        & {64.96$\pm$9.68}          \\
{LWF\cite{Li2018LearningWF}}                               & {13.71$\pm$7.94}         & {39.19$\pm$22.02}         & {6.22$\pm$3.78}          & {48.94$\pm$5.61}          & {12.57$\pm$0.69}         & {42.3$\pm$1.55}           & {0.07$\pm$0.10}           & {18.45$\pm$26.53}        \\
{MAS\cite{Aljundi2018MemoryAS}}                               & {21.91$\pm$1.94}         & {43.52$\pm$2.37}          & {30.56$\pm$8.57}         & {51.11$\pm$14.79}         & {17.36$\pm$2.33}         & {35.75$\pm$2.8}           & {23.36$\pm$8.88}        & {58.75$\pm$8.34}          \\
{SI\cite{Zenke2017ContinualLT}}                                & {21.56$\pm$2.25}         & {46.15$\pm$2.32}          & {31.46$\pm$9.71}         & {56.27$\pm$7.46}          & {17.53$\pm$0.99}         & {38.31$\pm$1.8}           & {31.16$\pm$5.44}         & {68.53$\pm$13.28}        \\ 
{DMC\cite{zhang2020class}}                               & {4.08$\pm$0.95}          & {19.20$\pm$2.80}          & {10.80$\pm$12.70}          & {33.79$\pm$10.13}          & {3.98$\pm$2.43}         & {19.22$\pm$3.53}          & {18.40$\pm$13.0}          & {48.46$\pm$19.14}           \\ \hline
{EEIL\cite{Castro2018EndtoEndIL}}                              & {5.65$\pm$0.68}          & {39.62$\pm$1.51}          & {1.16$\pm$0.8}           & {42.72$\pm$8.26}          & {5.25$\pm$0.56}          & {40.51$\pm$1.52}          & {15.33$\pm$8.74}          & {29.53$\pm$20.87}        \\
{ICARL\cite{Rebuffi2017iCaRLIC}}                             & {\textbf{0.61$\pm$0.4}}  & {11.79$\pm$1.24}          & {0.8$\pm$0.66}           & {18.96$\pm$4.28}          & {\textbf{0.75$\pm$0.33}} & {\textbf{11.97$\pm$0.63}} & {0.06$\pm$0.09}          & {{0.67$\pm$0.26}} \\
{BIC\cite{Wu2019LargeSI}}                               & {1.21$\pm$0.94}          & {\textbf{11.15$\pm$2.33}} & {\textbf{0.33$\pm$0.27}} & {{18.43$\pm$7.71}} & {1.78$\pm$0.44}          & {12.44$\pm$1.26}          & {\textbf{0$\pm$0}}       & {5.62$\pm$3.16}          \\
{DER++\cite{Buzzega2020DarkEF}}                               & {9.71$\pm$1.03}          & {40.85$\pm$0.87}          & {1.06$\pm$0.83}          & {30.22$\pm$5.38}          & {10.47$\pm$0.67}         & {41.93$\pm$1.77}          & {0.11$\pm$0.18}          & {2.18$\pm$0.30}           \\
 {HAL\cite{chaudhry2021using}}                               &  {9.63$\pm$4.10}          & {38.15$\pm$3.13}          & {1.57$\pm$2.21}          & {11.65$\pm$16.8}          &  {8.29$\pm$2.87}         &  {36.53$\pm$1.32}          & {45.80$\pm$0.19}          & {\textbf{0.02$\pm$0.04}}           \\
{\textbf{AGLA (Ours)}}               & {5.35$\pm$1.16} & {16.29$\pm$5.35} & {0.59$\pm$0.57} & {\textbf{17.13$\pm$1.65}}  & {3.78$\pm$0.26} & {15.63$\pm$6.77} & {0.06$\pm$0.04} & {0.89$\pm$0.37} \\ \hline
\end{tabular}
}
\caption{Average forgetting for continual learning benchmarks, averaged across 5 runs. $\ast$ 4 tasks only due to crash}
\label{tab:average_forgetting_all}
\end{table*}

\begin{table*}[ht]
\centering
\small
\begin{tabularx}{\textwidth}{lXXXXXXXXXX}
\hline
& \multicolumn{6}{c|}{{\textbf{Configuration}}}            &  \multicolumn{2}{c|}{\textbf{S-CIFAR-100}}                        & \multicolumn{2}{c}{\textbf{S-CIFAR-10}}                         \\ \cline{2-11} 
\multirow{2}{4em}{Code} &
Assr & Aug & R.Tr & ($\overline{w}_{i,k}$) & $\mathcal{L}$\tiny{{$_{DER++}$}} & $\mathcal{L}_{dist}$ & {Task IL}        & {Class IL}       & {Task IL}        & {Class IL}       \\ \hline
{A} & {}         & {\checkmark}            & {\checkmark}                & {\checkmark}            & {\checkmark}    & {\checkmark}     & {69.16}          & {46.26}          & {\textbf{96.42}} & {\textbf{82.96}} \\
{B} & {\checkmark}        & {}             & {\checkmark}                & {\checkmark}            & {\checkmark}    & {\checkmark}     & {\textbf{70.44}} & {45.83}          & {96.00}          & {80.50}          \\
{C} &  {\checkmark}        & {\checkmark}            & {}                 & {\checkmark}            & {\checkmark}    & {\checkmark}     & {68.52}          & {46.16}          & {96.00}          & {\textbf{82.94}} \\
{D} & {\checkmark}        & {\checkmark}            & {\checkmark}                & {}             & {\checkmark}    & {\checkmark}     & {69.50}          & {\textbf{46.82}} & {96.18}          & {82.74}          \\
{E} &  {\checkmark}        & {\checkmark}            & {\checkmark}                & {\checkmark}            & {}     & {\checkmark}     & {\textbf{70.01}} & {\textbf{47.61}} & {\textbf{96.28}} & {82.68}          \\
{F} & {\checkmark}        & {\checkmark}            & {\checkmark}                & {\checkmark}            & {\checkmark}    & {}      & {67.76}          & {43.98}          & {95.56}          & {78.78}          \\
{AGLA} & {\checkmark}        & {\checkmark}            & {\checkmark}                & {\checkmark}            & {\checkmark}    & {\checkmark}     & {\textbf{69.76}} & {\textbf{47.70}} & {\textbf{96.22}} & {\textbf{83.20}} \\\hline
\end{tabularx}
\caption{Classification result (accuracy) of AGLA based on various configuration}
\label{ablation_acc}
\end{table*}

\begin{table*}[ht]
\centering
\small
\begin{tabularx}{\textwidth}{lXXXXXXXXXX}
\hline
& \multicolumn{6}{c|}{{\textbf{Configuration}}}            &  \multicolumn{2}{c|}{\textbf{S-CIFAR-100}}                        & \multicolumn{2}{c}{\textbf{S-CIFAR-10}}                         \\ \cline{2-11} 
\multirow{2}{4em}{Code} &
Assr & Aug & R.Tr & ($\overline{w}_{i,k}$) & $\mathcal{L}$\tiny{{$_{DER++}$}} & $\mathcal{L}_{dist}$ & {Task IL}        & {Class IL}       & {Task IL}        & {Class IL}       \\ \hline
{A} & {}         & {\checkmark}            & {\checkmark}                & {\checkmark}            & {\checkmark}    & {\checkmark}     & {\textbf{4.04}} & {19.78}          & {\textbf{0.35}} & {{21.03}} \\
{B} & {\checkmark}        & {}             & {\checkmark}                & {\checkmark}            & {\checkmark}    & {\checkmark}     & {{5.30}} & {31.71}          & {0.68}          & {27.20}          \\
{C} & {\checkmark}        & {\checkmark}            & {}                 & {\checkmark}            & {\checkmark}    & {\checkmark}     & {\textbf{4.16}} & {19.00}          & {0.78}          & {\textbf{17.35}} \\
{D} & {\checkmark}        & {\checkmark}            & {\checkmark}                & {}             & {\checkmark}    & {\checkmark}     & {4.74}          & {\textbf{10.94}} & {\textbf{0.43}} & {20.85}          \\
{E} & {\checkmark}        & {\checkmark}            & {\checkmark}                & {\checkmark}            & {}     & {\checkmark}     & {\textbf{4.36}} & {\textbf{11.56}} & {{0.80}} & {\textbf{19.40}} \\
{F} & {\checkmark}        & {\checkmark}            & {\checkmark}                & {\checkmark}            & {\checkmark}    & {}      & {10.07}         & {36.44}          & {1.20}          & {29.83}          \\
{{AGLA}} & {\checkmark}        & {\checkmark}            & {\checkmark}                & {\checkmark}            & {\checkmark}    & {\checkmark}     & {{5.11}} & {\textbf{13.41}} & {\textbf{0.40}} & {\textbf{19.35}} \\ \hline
\end{tabularx}
\caption{Forgetting rate of AGLA based on various configuration}
\label{ablation_forg}
\end{table*}

\begin{table}[ht]
\centering
\small
\begin{tabular}{clcc}
\hline
{}                         & {}                         & \multicolumn{2}{c}{{S-CIFAR-100 Accuracy}}                                           \\ \cline{2-4} 
{Memory} & {Method} & {Task IL}        & {Class IL}           \\ \hline
{}                         & {EEIL\cite{Castro2018EndtoEndIL}}                     & {63.88}          & {33.43}                   \\
{}                         & {ICARL\cite{Rebuffi2017iCaRLIC}}                    & {63.94}          & {40.07}                  \\
{2000}                     & {BIC\cite{Wu2019LargeSI}}                      & {{67.41}} & {{42.33}}        \\
{}                         & {DER++\cite{Buzzega2020DarkEF}}                      & {62.53}          & {34.33}                \\
{}                         &  {HAL\cite{chaudhry2021using}}                      &  {55.92}          & {28.02}                \\
{}                         & \textbf{AGLA (Ours)}  & {\textbf{67.60}}             & {\textbf{43.50}}          \\ \hline
{}                         & {EEIL\cite{Castro2018EndtoEndIL}}                     & {65.7}           & {36.94}                   \\
{}                         & {ICARL\cite{Rebuffi2017iCaRLIC}}                    & {65.08} & {42.01}          \\
{3000}                     & {BIC\cite{Wu2019LargeSI}}                      & {{68.21}}          & {{44.1}}                    \\
{}                         & {DER++\cite{Buzzega2020DarkEF}}                      & {64.26}          & {36.95}                  \\
{}                         & {HAL\cite{chaudhry2021using}}                      & {57.23}          & {30.75}                \\
{}                         & \textbf{AGLA (Ours)}     & {\textbf{69.07}}          & {\textbf{45.64}}        \\ \hline
{}                         & {EEIL\cite{Castro2018EndtoEndIL}}                     & {66.12}          & {38.54}                 \\
{}                         & {ICARL\cite{Rebuffi2017iCaRLIC}}                    & {65.34}          & {42.13}                 \\
{4000}                     & {BIC\cite{Wu2019LargeSI}}                      & {67.26}          & {44.28}               \\
{}                         & {DER++\cite{Buzzega2020DarkEF}}                      & {65.06}          & {39.43}              \\
{}                         & {HAL\cite{chaudhry2021using}}                      &  {57.77}          & {32.19}                \\
{}                         & \textbf{AGLA (Ours)}      & {\textbf{69.34}} & {\textbf{46.19}} \\ \hline
{}                         & {EEIL\cite{Castro2018EndtoEndIL}}                     & {66.74}          & {40.18}               \\
{}                         & {ICARL\cite{Rebuffi2017iCaRLIC}}                    & {66.09}          & {43.03}                 \\
{5000}                     & {BIC\cite{Wu2019LargeSI}}                      & {67.84}          & {44.8}                   \\
{}                         & {DER++\cite{Buzzega2020DarkEF}}                      & {66.57}          & {41.33}               \\
{}                         &  {HAL\cite{chaudhry2021using}}                      &  {57.76}          &  {32.32}                \\
{}                         & \textbf{AGLA (Ours)}     & {\textbf{69.34}} & {\textbf{47.70}}   \\ \hline
\end{tabular}
\caption{Classification result (accuracy) of the benchmarks based on memory size}
\label{memory_analysis_acc}
\end{table}

\begin{table}[ht]
\centering
\small
\begin{tabular}{clcc}
\hline
{}                         & {}                         & \multicolumn{2}{c}{{S-CIFAR-100 Forgetting rate}} \\ \cline{2-4} 
{Memory} & {Method} & {Task IL}       & {Class IL}  \\ \hline
{}                         & {EEIL\cite{Castro2018EndtoEndIL}}                     & {8.14}          & {49.9}                \\
{}                         & {ICARL\cite{Rebuffi2017iCaRLIC}}                    & {\textbf{2.47}} & {15.71}                 \\
{2000}                     & {BIC\cite{Wu2019LargeSI}}                      & {2.96}          & {\textbf{15.54}} \\
{}                         & {DER++\cite{Buzzega2020DarkEF}}                      & {14.99}         & {52.81}                 \\
{}                         &  {HAL\cite{chaudhry2021using}}                      &  {10.72}          &  {47.46}                \\
{}                         & \textbf{AGLA (Ours)}     &  {8.37}          &  {31.37}                  \\ \hline
{}                         & {EEIL\cite{Castro2018EndtoEndIL}}                     & {7.78}          & {46.44}               \\
{}                         & {ICARL\cite{Rebuffi2017iCaRLIC}}                    & {\textbf{2.52}} & {15.32}         \\
{3000}                     & {BIC\cite{Wu2019LargeSI}}                      & {2.58}          & {\textbf{13.99}}    \\
{}                         & {DER++\cite{Buzzega2020DarkEF}}                      & {11.8}          & {48.23}                  \\
{}                         &  {HAL\cite{chaudhry2021using}}                      &  {9.02}          &  {41.80}                \\

{}                         & \textbf{AGLA (Ours)}     &  {5.21}          &  {19.58}           \\ \hline
{}                         & {EEIL\cite{Castro2018EndtoEndIL}}                     & {5.84}          & {42.58}               \\
{}                         & {ICARL\cite{Rebuffi2017iCaRLIC}}                    & {\textbf{0.69}} & {\textbf{12.59}}  \\
{4000}                     & {BIC\cite{Wu2019LargeSI}}                      & {2.23}          & {13.69}                 \\
{}                         & {DER++\cite{Buzzega2020DarkEF}}                      & {10.18}         & {42.94}                  \\
{}                         & {HAL\cite{chaudhry2021using}}                      &  {9.02}          &  {40.32}                \\
{}                         & \textbf{AGLA (Ours)}     &  {5.66}          &  {24.88} \\ \hline
{}                         & {EEIL\cite{Castro2018EndtoEndIL}}                     & {5.71}          & {40.14}                 \\
{}                         & {ICARL\cite{Rebuffi2017iCaRLIC}}                    & {\textbf{0.31}} & {\textbf{10.73}}      \\
{5000}                     & {BIC\cite{Wu2019LargeSI}}                      & {1.87}          & {13.32}                \\
{}                         & {DER++\cite{Buzzega2020DarkEF}}                      & {8.33}          & {39.87}               \\
{}                         &  {HAL\cite{chaudhry2021using}}                      &  {8.18}          &  {37.20}                \\
{}                         & \textbf{AGLA (Ours)}     &  {5.11}          &  {14.54}     \\ \hline
\end{tabular}
\caption{Forgetting rate of the benchmarks based on memory size}
\label{memory_analysis_forg}
\end{table}

\section{Experiments}
The advantage of AGLA is numerically validated in both class-incremental and task-incremental learning problems. In addition, ablation study and memory analysis are offered to analyze the contribution of each component and the memory size to the final performance. Our experiments are conducted by five independent runs using different random seeds where the numerical results are averaged and reported in Table \ref{tab:average_accuracy_all} and Table \ref{tab:average_forgetting_all}. Two performance metrics, average accuracy and forgetting measure \cite{Chaudhry2019OnTE}, are applied to measure the learning performances. The source codes and other supporting data of AGLA including raw numerical results are made public in \url{https://github.com/anwarmaxsum/AGLA} for further study. 
The models are evaluated in two continual learning setting i.e. task incremental learning (Task IL) and class incremental learning (Class IL). \textbf{Task IL} is a problem setting where a model is trained to a sequence of tasks $\mathcal{T}_1,\mathcal{T}_2,...,\mathcal{T}_{K}, k\in\{1,...,K\}$, each task $\mathcal{T}_{k}$ has its training set  $T_k=\{x_n,y_n,id_n\}_{n=1}^{N_k}$, $x_n$ represents an input sample and $y_{n}\in\mathcal{Y}$ denotes its true class label, $id_n$ stands for the task id. $K, N_k$ respectively denote the number of tasks and the size of $k-th$ task. \textbf{Class IL} is a similar setting with Task IL but the task id $id_n$ is not visible to the model.

\subsection{Datasets}
Four datasets examine the advantage of AGLA: split MNIST (SMNIST) \cite{LeCun1998GradientbasedLA,Zenke2017ContinualLT}, split CIFAR10 (SCIFAR10) \cite{Krizhevsky2009LearningML,Hu2021ContinualLB}, split CIFAR100 (SCIFAR100) \cite{Krizhevsky2009LearningML,vandeVen2020BraininspiredRF} and Split Mini-ImageNet (SMImageNet) \cite{Vinyals2016MatchingNF,Pham2021ContextualTN}. SMNIST, SCIFAR10 feature the task-incremental and class-incremental learning problems with 5 tasks where each task contains two disjoint classes whereas SCIFAR100 and SMImageNet also present the two problems with 10 tasks where each task features 10 disjoint classes. Note that no task-IDs and the single-head structure are applied in the class-incremental learning problem while the task-incremental learning problem benefits from the presence of task IDs and the multi-head network structure. Table \ref{tab:problems} tabulates our experimental setting. The hyperparameters setting for the consolidated algorithms is presented in Table 3. 

\subsection{Baseline Algorithms}
AGLA is compared with eight algorithms: EWC \cite{Kirkpatrick2017OvercomingCF}, SI \cite{Zenke2017ContinualLT}, LWF \cite{Li2018LearningWF}, MAS \cite{Aljundi2018MemoryAS}, iCaRL \cite{Rebuffi2017iCaRLIC}, DER++ \cite{Buzzega2020DarkEF}, EEIL \cite{Castro2018EndtoEndIL}, BIC \cite{Wu2019LargeSI}, DMC \cite{zhang2020class} and HAL \cite{chaudhry2021using} . In addition, the lower (fine-tuning) and upper bounds (joint) are provided where the naive SGD depicts the lower bound case and the joint training approach presents the upper bound case. Comparison is done in the same computational environments, eight NVIDIA DGX A100 GPUs with 40 GB each, under the FACIL framework \cite{masana2020class} to ensure fair comparison.  

\subsection{Experimental Setup}
For all problems, ResNet18 is deployed without the pretraining process. A grid search approach is applied to all consolidated algorithms where detailed hyper-parameters are provided in Table 3. On the other hand, the assessor is formed as the LSTM network: ResNet18 feature extractor, two LSTM layers and a single fully connected layer where the number of nodes in each layer is set as 64. The memory size of the memory-based approaches, AGLA, BIC, DER++, EEIL and ICaRL, is set equally to 5000 samples for all problems but the memory analysis is offered here as well. The image size of the SMImageNet is resized to 32 by 32 from the original size of 84 by 84. The hyper-parameters of consolidated algorithms are listed in the Table VII. Note that AGLA only carries a single hyper-parameter, memory size, common to other memory-based approaches. Other hyper-parameters are shared by all benchmarked algorithms.  

\subsection{Numerical Results}
From Table \ref{tab:average_accuracy_all}, it is shown that AGLA outperforms other algorithms in all four problems with noticeable gaps. AGLA produces better accuracy than BIC having the second highest accuracy by about $2-3\%$ in the SCIFAR100 problem for both the task-incremental and class-incremental learning problems. This finding confirms the advantage of the meta-trained assessor in guiding the learning process of the base model. The assessor enables a soft sample selection mechanism assigning high weights only to positive samples and a dynamic weighting mechanism of loss functions controlling interactions of the three loss functions. Note that BIC features the bias-correction layer to handle the class-imbalanced problem. The class-imbalanced problem is addressed in AGLA with the compensated over-sampling technique which does not call for an additional training stage as per BIC. This is done by simply augmenting the old samples in the memory with corrections to prevent out of distributions. The same pattern is observed in the SCIFAR10 problem where AGLA delivers the highest accuracy with about $1.2\%$ gap to DER++ in the task-incremental learning problem and $4\%$ gap to ICARL in the class-incremental learning problem. 

AGLA is also superior to other algorithms in the SMImagenet problem. It beats the BIC by $1.5\%$ margin in both the task-incremental and class-incremental learning problems. Once again, this aspect confirms the assessor guided learning process of AGLA achieving better tradeoff between the plasticity and the stability than other algorithms. The use of the meta-trained assessor also performs a selection of suitable learning strategy of data samples. It controls how much new concepts and past concepts are to be accepted. On the other hand, AGLA also outperforms other consolidated algorithms with a tiny margin in the task-incremental learning and class-incremental learning of the SMNIST problem. Note that the SMNIST problem is considered as an easy problem for continual learning algorithms, i.e., most algorithms produce decent results in this problem. The performance of AGLA is also stable across the four problems. That is, it does not suffer from any sudden performance losses, i.e., this occurs for BIC (SCIFAR10) and DER++ (SMINIIMAGENET). Another important finding is seen in the fact that the memory-based approaches are better than the regularization-based approaches. This fact becomes obvious in the class-incremental learning problem. Nonetheless, the regularization-based approach is simple to implement and faster than the memory-based approach. Table \ref{tab:average_forgetting_all} presents the average forgetting of all consolidated algorithms. It is perceived that AGLA produces comparable average forgetting to other memory-based approaches where it mostly occupies the first or second place and still far better than the regularization-based approaches. This should not hinder the advantage of AGLA because the average forgetting is referred to only if the accuracy of compared algorithms is on par, i.e., AGLA beats other algorithms in all four cases. Low average forgetting score might occur in the case of poor average accuracy.   



\begin{figure}
\includegraphics[width=1.0\textwidth]{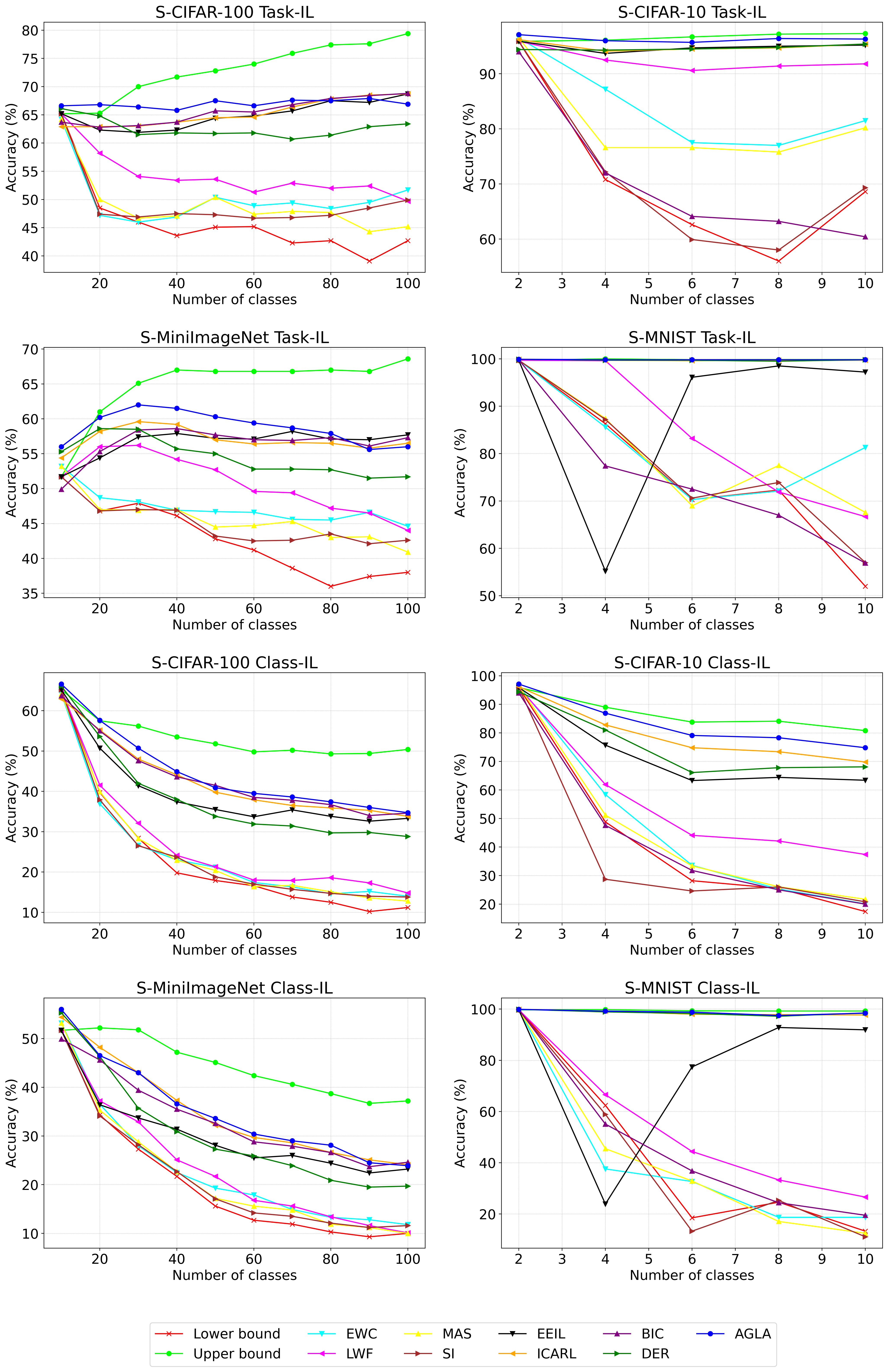}
\caption{Classification Accuracy of Consolidated Algorithms across All Datasets.}
\label{Fig:Average Accuracy on 4 dataset}
\end{figure}

\begin{figure}
\includegraphics[width=1.0\textwidth]{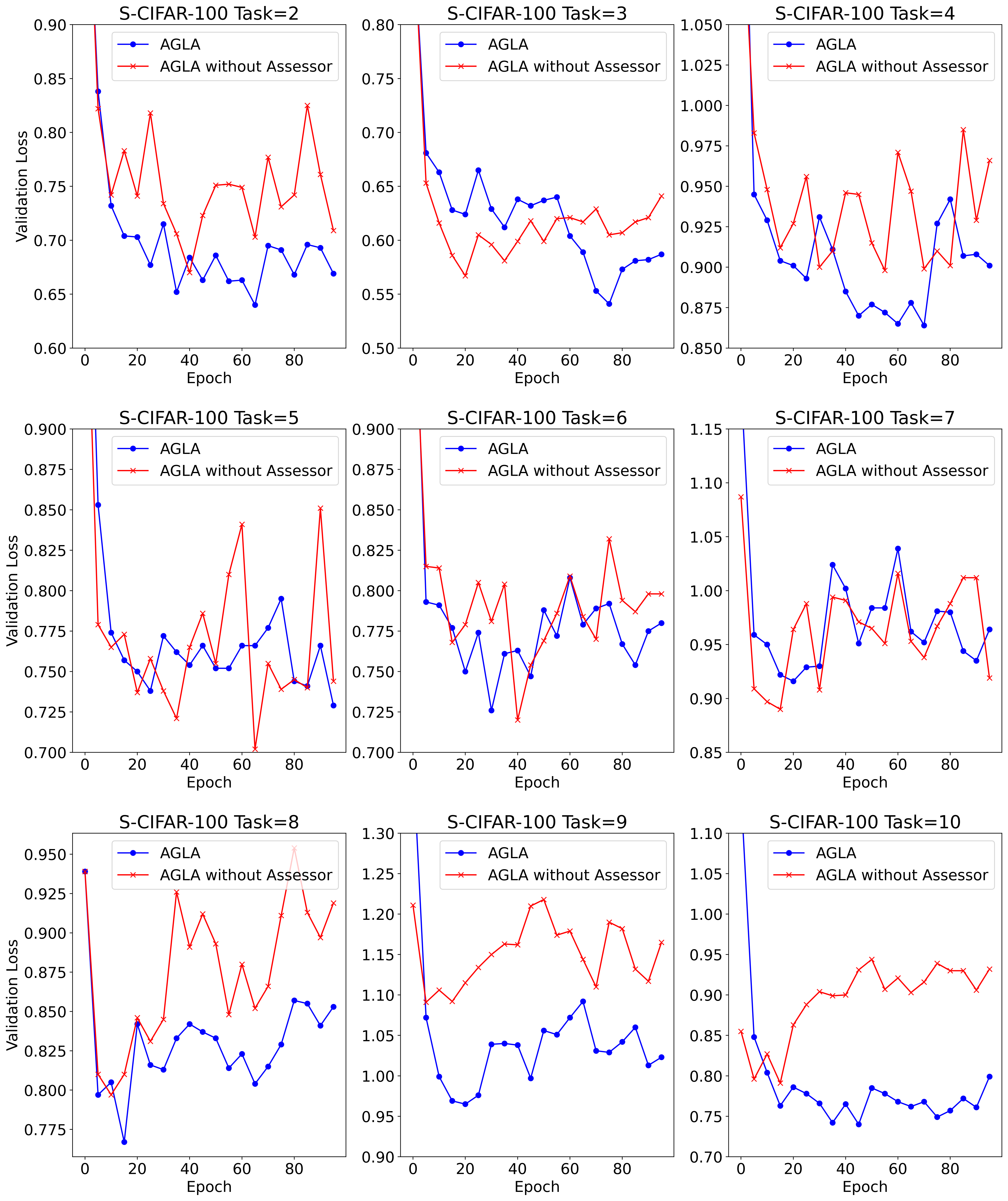}
\caption{The trace of losses for AGLA with or w/o Assessor in the S-CIFAR-100 problem.}
\label{Fig:AGLA vs AGLA Without Assessor}
\end{figure}

\begin{figure}
\includegraphics[width=1.0\textwidth]{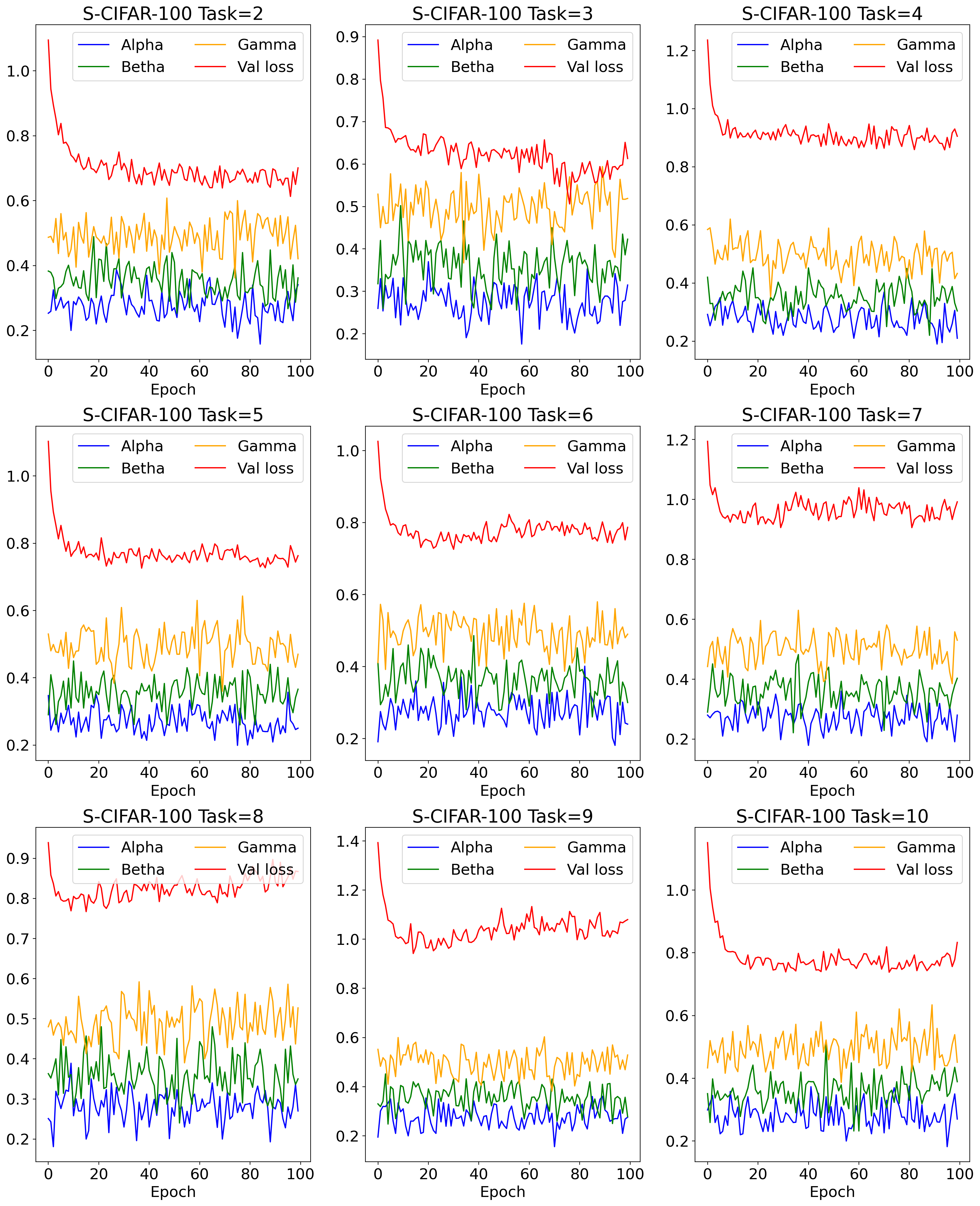}
\caption{The trace of weighting coefficients and loss for AGLA in the S-CIFAR-100 problem.}
\label{Fig:Coefficients and Loss of AGLA}
\end{figure}

\subsection{Ablation Study}
This section demonstrates the contribution of each learning component of AGLA. Five aspects are studied here: 1) the learning performance of AGLA without the presence of assessor is presented - configuration (A);  2) the learning performance of AGLA without the data augmentation strategy is demonstrated - configuration (B). That is, no oversampling strategy is performed; 3) the learning performance of AGLA without the random transformation strategy to construct the validation set of the meta-learning phase is depicted - configuration (C). This implies the use of the training set in updating the assessor; 4) the learning performance of AGLA without the out-of-distribution compensation is depicted - configuration (D). This implies that the augmented data is not guaranteed to be within the original samples distribution; 5) the learning performance of AGLA without the DER loss is shown - configuration (E). This implies the application of the two loss functions only, the cross entropy loss function and the distillation loss function; 6) the learning performance of AGLA without the distillation loss function is presented - configuration (F). This implies the use of the cross entropy loss function and the DER loss function. All numerical results are tabulated in Table \ref{ablation_acc} and in Table \ref{ablation_forg} where they are produced using the SCIFAR10/100 problems in both the class-incremental learning problem and the task-incremental learning problem. Note that we are only able to report numerical results of a single run for the ablation study due to limited accesses of computational resources.

The absence of the assessor as depicted in configuration (A) leads to significant performance deterioration in all cases with up to $2\%$ drop in accuracy and $7\%$ drop in average forgetting. This finding confirms the importance of assessor in balancing between the plasticity and the stability. The three loss functions of AGLA contribute positively to the final performance. The absence of the distillation loss function in configuration (F) or the DER loss function in configuration (E) cause major performance drop, i.e., up to $15\%$ for configuration (F) and up to $1\%$ for configuration (E). The DER loss function and the distillation loss function protect against the catastrophic forgetting. This finding is further borne out by the major increase of average forgetting. Table \ref{ablation_forg} shows that the absence of distillation loss and DER loss increases the forgetting rate up to $23\%$ for configuration (F) and $2\%$ for configuration (E). The over-sampling procedure plays vital role where configuration (B) shows performance degradation in accuracy (up to $3\%$) and average forgetting (up to $18\%$). This mechanism is designed to cope with the class imbalanced problem of the memory-based approach. The random transformation technique to construct the validation set for the assessor training process improves model's generalization, i.e., configuration (C) results in noticeable loss in performance (up to $2\%$) and increasing forgetting rate (up to $6\%$). This aspect demonstrates that the meta-training strategy should be carried out using different training and validation sets. Last but not least, the compensation strategy also plays important role to the proposed method i.e. configuration (D) leads to drops in accuracy by up to $1\%$ and increasing in forgetting rate by up to $2\%$. This finding shows that the augmented samples should be protected to be within the original samples distribution. 

\subsection{Memory Analysis}
This section depicts the performance of AGLA and other consolidated memory-based approaches with different memory sizes in the SCIFAR100 problem under both class-incremental and task-incremental configurations. Specifically, the memory size is set to 2000, 3000, 4000 respectively where consolidated numerical results are provided in Table \ref{memory_analysis_acc} and Table \ref{memory_analysis_forg}. As with the ablation study, the memory analysis is performed under a single run due to limited accesses of computational resources. However, this should not bias our finding in the ablation study and the memory analysis since AGLA is superior to all baseline algorithms in all cases in our main results reported in Table \ref{tab:average_accuracy_all} and \ref{tab:average_forgetting_all} where five runs are committed under different random seeds.  

It is obvious that AGLA still maintains superior performance compared to other memory-based approaches under smaller memory sizes (2000, 3000, 4000) than that reported in Table \ref{tab:average_accuracy_all} and \ref{tab:average_forgetting_all} (5000) with clear margins in the range of $1-3\%$. The performances of AGLA are relatively stable under different memory budgets where performance drops due to reductions of memory sizes are less than $4\%$. Although ICARL and BIC produced smaller average forgetting than AGLA, their accuracy are far worse than AGLA. This finding confirms the advantage of the compensated over-sampling approach to cope with the severe class imbalanced problem without any extra parameters. It is also observed that the performance drops due to the memory constraint are higher in the class-incremental learning problem than in the task-incremental learning problem because of the absence of task IDs and single-head classifiers. 

\subsection{Performance Evaluations per Task}
The evolution of classification accuracy of all consolidated algorithms in all datasets in both the class-incremental and task-incremental learning problems in one of runs are pictorially illustrated in Fig. \ref{Fig:Average Accuracy on 4 dataset} where the advantage of AGLA is clearly demonstrated. For the class-IL problems, AGLA produces higher classification accuracy in all tasks across all datasets than its counterparts. 

For task-IL problem, AGLA outperforms other algorithms across all tasks in all datasets. That is, it returns the highest classification accuracy per task than other algorithms. The positive contribution of the assessor-guided continual learning is portrayed here where it beats the DER++ algorithm across all tasks in all datasets regardless of the task-IL or class-IL problem. Note that AGLA shares similar loss functions as DER++ except for the application of the distillation loss function and the assessor and the meta-weighting strategy.  

\subsection{Assessor vs w/o Assessor}
This subsection confirms the positive roles of the assessor in guiding the continual learning process of AGLA where Fig. \ref{Fig:AGLA vs AGLA Without Assessor} visualizes the trace of AGLA's losses in the SCIFAR100 problem with and w/o the assessor under the class-IL setting in one of the experiments. Clearly, the use of assessor to perform the soft-weighting mechanism of data samples and the selection of appropriate learning strategies result in faster convergence and lower losses than with the absence of assessor in most tasks. Importantly, AGLA remains stable across all tasks with the assessor-guided learning process, i.e., the losses signify decreasing trends across all tasks. This fact implies the success of our meta-learning-based bi-level optimization approach to fine-tune both the base network and the assessor.

\subsection{Evolution of Meta-weights}
Figure \ref{Fig:Coefficients and Loss of AGLA} shows the plots visualizing the evolutions of meta-weights ($\alpha$, $\beta$, $\gamma$) in CIFAR-100 dataset. The figure shows that the value of $\gamma$ and $\beta$ that correspond to distillation loss and dark experience replay loss respectively are higher than the value of $\alpha$ that corresponds to cross-entropy loss. It shows that the proposed method needs higher weights for the memory samples than the current samples. This is in line with the fact that the number of memory samples is less than the current task samples, so it needs higher weights in current task training to handle class-imbalance problems. The figure also shows that the dynamic evolution of the three weights leads to decreasing trends of validation losses.

\subsection{Complexity Analysis}
This sub-section discusses the complexity analysis of the proposed method. Suppose that   $N$ is the total samples of a dataset, $k \in\{1,...,K\}$ is the index of task, $N_k$ is the number of samples on task $k$, where satisfy $\mathcal{\sum}_{k=1}^{K} N_k = N$, $e$ is the number of epoch for networks training, $M_k$ is the size of memory on task $k$ that satisfy $\mathcal{\sum}_{k=1}^{K} M_k < N$. Following the pseudo-code in Algorithm 1, there are several processes one each task i.e. augmentation (Aug), union of train data and memory (Union), random transformation (R.Trans), assessor and base networks update (N.Update), memory sampling (M.Sampling) and memory update (M.Update). Please note that augmentation and random transformation are done in a few operations $(c < 10)$. Let $C$ denote the complexity of a process. Following the Algorithms 1, the complexity of the proposed method can be written as the following equations:

\begin{equation}
    \begin{split}
        \mathcal{C}(AGLA) & =\mathcal{\sum}_{k=1}^{K} (\mathcal{C}(Aug) + \mathcal{C}(Union) + \mathcal{C}(R.Trans) + \\ 
        & \mathcal{C}(N.update) + \mathcal{C}(M.Samping) + \mathcal{C}(M.Update))
    \end{split}
\end{equation}

\begin{equation}
    \begin{split}
        \mathcal{C}(AGLA) & =\mathcal{\sum}_{k=1}^{K} ( (c.N_k) + max(N_k,M_k) + (c.(N_k+M_k)) + \\ 
        & (2 . e. (N_k+M_k)) + (N_k+M_k) + max(N_k+M_k, M_{k+1}))
    \end{split}
\end{equation}

\begin{equation}
    \begin{split}
        \mathcal{C}(AGLA) & =( (c. \mathcal{\sum}_{k=1}^{K} N_k) + max( \mathcal{\sum}_{k=1}^{K} N_k, \mathcal{\sum}_{k=1}^{K} M_k) + \\
        & (c.(\mathcal{\sum}_{k=1}^{K} N_k+ \mathcal{\sum}_{k=1}^{K} M_k)) + \\ 
        & (2 . e. (\mathcal{\sum}_{k=1}^{K} N_k+ \mathcal{\sum}_{k=1}^{K} M_k)) + \\
        & (\mathcal{\sum}_{k=1}^{K} N_k+\mathcal{\sum}_{k=1}^{K} M_k) + \\
        & max( \mathcal{\sum}_{k=1}^{K} N_k + \mathcal{\sum}_{k=1}^{K} M_k, \mathcal{\sum}_{k=1}^{K} M_{k+1}))
    \end{split}
\end{equation}

Since $\mathcal{\sum}_{k=1}^{K} N_k = N$, $\mathcal{\sum}_{k=1}^{K} M_k < N$, and $\mathcal{\sum}_{k=1}^{K} M_{k+1} < N$. then the complexity of AGLA can be derived to:

\begin{equation}
    \begin{split}
        \mathcal{C}(AGLA) & \le ( (c.N) + N + (c.(N+N)) + \\ 
        & (2 . e. (N+N)) + (N+N) + (N+N))
    \end{split}
\end{equation}
\begin{equation}
    \begin{split}
        \mathcal{C}(AGLA) & \le (cN + N + 2cN + 4eN + 2N + N)
    \end{split}
\end{equation}
\begin{equation}
    \begin{split}
        \mathcal{C}(AGLA) & \le (4N + 3cN + 4eN)
    \end{split}
\end{equation}

\begin{equation}
    \begin{split}
        \mathcal{C}(AGLA) & = (O(N) + O(cN) + O(eN))
    \end{split}
\end{equation}

Considering that $c$ is a small number $( < 10)$, then the complexity of AGLA can be derived to:

\begin{equation}\label{}
    \begin{split}
        \mathcal{C}(AGLA) & = (O(N) + O(N) + O(eN)) \\
        & = (O(N) + O(eN)) \\
        & = O(eN) \\
    \end{split}
\end{equation}

The equation above concludes that the complexity of the proposed method is $O(eN)$, where N is the total instance of the data across all tasks, and e is the number of epochs for network training. In case that the epoch is set as constant e.g. 100, then the complexity of the proposed method will be $O(N)$.

\section{Conclusions}
This paper proposes an assessor-guided learning approach (AGLA) for continual learning where it puts forward a sequence-aware assessor to guide the learning process of the base learner. The assessor performs a soft-weighting mechanism controlling the influence of a data sample and the interaction of loss functions. It is made possible by forming the loss function of the base learner as the meta-weighted combination of the cross entropy loss function, the DER loss function and the distillation loss function. The underlying objective is to arrive at proper tradeoff between the plasticity and the stability because not only positive samples are selected, but also suitable learning strategies are chosen to every sample, i.e., the assessor balances the stability and the plasticity for every sample. Another major contribution is perceived in the proposal of compensated over-sampling (COS) to address the class-imbalanced problem on continual learning where corrections are carried out while augmenting data samples to avoid the effect of out of distribution augmented samples. Rigorous numerical results have been carried out using four popular continual learning problems in both the task-incremental and class-incremental learning problems. AGLA has been compared with ten recently published algorithms under the same computational environments where AGLA demonstrates improved and stable accuracy across all four problems in both the task-incremental learning problems and the class-incremental learning problems with noticeable margins. In realm of average forgetting, AGLA delivers comparable performances to those of memory-based approaches where it mostly occupies the second lowest average forgetting index and better than those of regularization-based approaches. Our ablation study demonstrates the advantage of each learning component of AGLA where the absence of one component results in major performance drops. AGLA also maintains decent performances with tiny memory sizes as indicated by our memory analysis. Existing continual learning approaches including AGLA function with a large number of samples and over-fits quickly in the case of few samples in each task. 

Our future work is directed to answer the few-shot continual learning problems handling a limited sample problem. The few-shot continual learning is a challenge since it is difficult to achieve plasticity while maintaining stability with few-shot samples. Furthermore, the replay methods that save previous task samples as memory can not be applied since they will be the same as joint training. Our future work is also directed to unsupervised few-shot continual learning problems to address unavailable labels in the dynamic environment. In real-world applications, a dataset is feasible to collect, but the labels (annotations) are hardly available. It needs expensive human effort to annotate the dataset. Last, but not least, our work is also directed to address federated continual learning problems to handle the data privacy constraint. The problems simulate there are many agents e.g. institutions that conduct continual learning in their respective environments to collaborate with each other but without sharing their private data.

\section{Acknowledgement}
\noindent This work is financially supported by the UniSA's start-up grant. The third authors acknowledges the support of the COMET-K2 Center of the Linz Center of Mechatronics (LCM) funded by the Austrian federal government and the federal state of Upper Austria.

\bibliography{myreference}
\end{document}